\documentclass[10pt,twocolumn,letterpaper]{article}

\usepackage{amsmath,amsfonts,bm}

\def\eqref#1{equation~\ref{#1}}

\def\1{\bm{1}}

\def\vp{{\bm{p}}}

\def\vw{{\bm{w}}}
\def\vx{{\bm{x}}}
\def\vy{{\bm{y}}}

\DeclareMathAlphabet{\mathsfit}{\encodingdefault}{\sfdefault}{m}{sl}
\SetMathAlphabet{\mathsfit}{bold}{\encodingdefault}{\sfdefault}{bx}{n}

\usepackage{cvpr}              

\usepackage{graphicx,adjustbox}
\usepackage{url}
\usepackage{amssymb}
\usepackage[linesnumbered,ruled,vlined]{algorithm2e}
\SetAlFnt{\small}
\SetKwComment{Comment}{/* }{ */}
\SetKwInput{kwInput}{Input}
\SetKwInput{kwOutput}{Output}
\RestyleAlgo{ruled}
\usepackage{amsmath}
\usepackage{array}
\usepackage{multirow}
\usepackage{booktabs}
\usepackage[table]{xcolor}
\usepackage[normalem]{ulem}
\useunder{\uline}{\ul}{}
\usepackage{float, wrapfig}
\usepackage{resizegather}
\usepackage{enumitem}
\usepackage{diagbox}

\usepackage[font=small,skip=0pt]{caption}
\usepackage[accsupp]{axessibility}

\usepackage{amssymb}

\usepackage[pagebackref,breaklinks,colorlinks]{hyperref}
\usepackage[capitalize]{cleveref}
\crefname{section}{Sec.}{Secs.}
\Crefname{section}{Section}{Sections}
\Crefname{table}{Table}{Tables}
\crefname{table}{Tab.}{Tabs.}

\def\cvprPaperID{15712} 
\def\confName{CVPR}
\def\confYear{2025}
\begin{document}


\title{Enhancing Dataset Distillation via Non-Critical Region Refinement}

\author{Minh-Tuan Tran$^1$, Trung Le$^1$, Xuan-May Le$^2$, Thanh-Toan Do$^1$, Dinh Phung$^1$\\\
$^1$Monash University,  $^2$The University of Melbourne\\
{\tt\small \{tuan.tran7,trunglm,toan.do,dinh.phung\}@monash.edu} \\ \tt\small xuanmay.le@student.unimelb.edu.au}


\maketitle

\begin{abstract}
Dataset distillation has become a popular method for compressing large datasets into smaller, more efficient representations while preserving critical information for model training. Data features are broadly categorized into two types: instance-specific features, which capture unique, fine-grained details of individual examples, and class-general features, which represent shared, broad patterns across a class. However, previous approaches often struggle to balance these features—some focus solely on class-general patterns, neglecting finer instance details, while others prioritize instance-specific features, overlooking the shared characteristics essential for class-level understanding. In this paper, we introduce the Non-Critical Region Refinement Dataset Distillation (NRR-DD) method, which preserves instance-specific details and fine-grained regions in synthetic data while enriching non-critical regions with class-general information. This approach enables models to leverage all pixel information, capturing both feature types and enhancing overall performance. Additionally, we present Distance-Based Representative (DBR) knowledge transfer, which eliminates the need for soft labels in training by relying on the distance between synthetic data predictions and one-hot encoded labels. Experimental results show that NRR-DD achieves state-of-the-art performance on both small- and large-scale datasets. Furthermore, by storing only two distances per instance, our method delivers comparable results across various settings. The code is available at \url{https://github.com/tmtuan1307/NRR-DD}.


\end{abstract}

\section{Introduction}

Dataset distillation, also known as dataset condensation \cite{dd, ddgm, ddtm, rded, sre2l}, has gained considerable attention as an effective method for compressing large datasets into smaller, more efficient representations, while preserving essential information critical for model training. By generating compact, high-quality data representations, dataset distillation reduces both storage requirements and computational costs associated with training on full-sized datasets \cite{vit, resnet, sbert}. This compression is particularly valuable in resource-constrained environments, such as edge devices or memory-limited systems, where training on large datasets is often impractical \cite{ocl, rfl, lander}. The primary goal of dataset distillation is to achieve high model performance while drastically reducing the amount of data that needs to be stored and processed \cite{dd}. During this process, synthetic data instances are generated to capture the key properties of the original data, enabling models to generalize effectively using only a fraction of the dataset \cite{ddgm, ddtm}.

In recent years, several methods have been proposed to address this task, including Gradient Matching \cite{ddgm, ddgm2}, Distribution Matching \cite{idm, dddm2}, Trajectory Matching \cite{ddtm, ddtm2}, and more recent approaches for large-scale datasets \cite{tesla, sre2l, rded}. However, designing an efficient distillation method remains a challenge, as it must capture both class-general features, which represent shared patterns, and instance-specific features, which highlight unique, fine-grained details. Previous methods often fall short by emphasizing one of these feature types over the other. Approaches focusing on class-general features \cite{ddgm, ddtm, idm, tesla, sre2l} risk losing crucial instance-specific information, which hinders fine-grained generalization. Conversely, methods that prioritize instance-specific features \cite{rded, mmdiff} may neglect broader class patterns, leading to suboptimal class-level representation.
To address these challenges, we introduce the Non-Critical Region Refinement Dataset Distillation (NRR-DD) method, which consists of three key stages:

\begin{enumerate}[noitemsep, nolistsep] 
\item[(i)] \textbf{Critical-based Initial Data Discovery}: This stage involves selecting diverse and significant patches from the original dataset, which are then combined to capture instance-specific features. 
\item[(ii)] \textbf{Non-Critical Region Refinement (NRR)}: In this stage, we apply Class Activation Mapping (CAM) \cite{cam, gradcam} to identify critical and non-critical regions in the images. The model preserves the critical regions, which contain fine-grained, instance-specific features, while refining the non-critical regions with more class-general information. By balancing these two feature types, NRR enhances the dataset’s comprehensiveness, improving both generalization and performance. 
\item[(iii)] \textbf{Knowledge Transfer via Relabeling}: After training, the synthetic images are relabeled and used to transfer knowledge to a student model. 
\end{enumerate}

Additionally, recent research \cite{sre2l, tesla, rded} has highlighted the importance of soft labels generated by a pretrained model in enhancing dataset distillation performance, particularly for large-scale, high-resolution datasets like ImageNet1k. However, this approach incurs substantial memory overhead; for instance, with 200 images per class (IPC) in ImageNet1k, it can require over 120 GB of storage \cite{sre2l}.

\begin{figure}[t]
\begin{center}
\includegraphics[width=\linewidth]{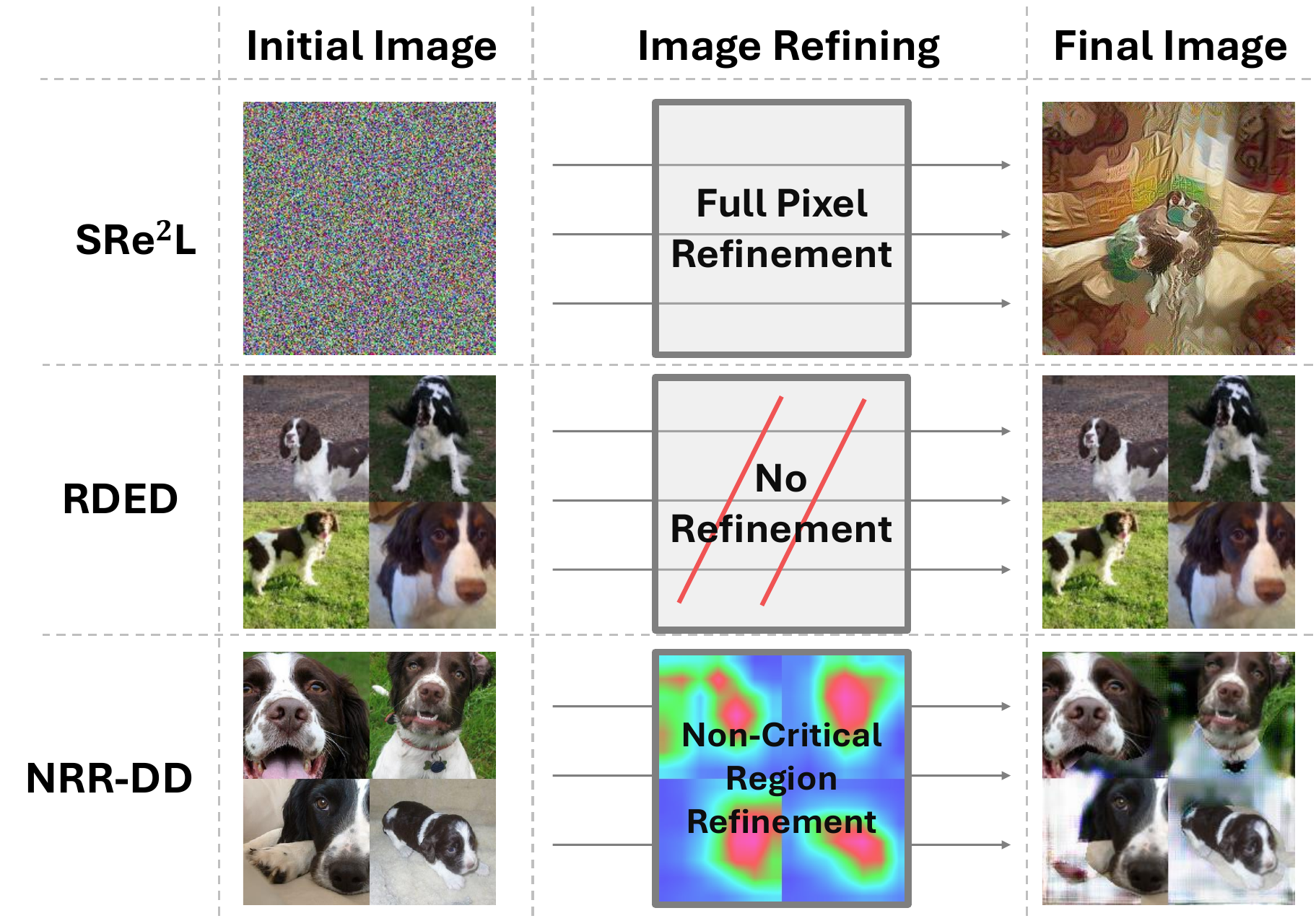}
\end{center}
\caption{Comparison of our method with two popular frameworks, SRe$^2$L \cite{sre2l} and RDED \cite{rded}, for generating synthetic datasets. RDED selects high-confidence, easily classifiable images, while our method focuses on low-confidence, harder-to-classify samples, which helps reduce overfitting and improve model accuracy. Additionally, RDED targets instance-specific features without refinement, and SRe$^2$L updates all pixels to capture class-general features, often at the expense of fine details. In contrast, our NRR-DD method preserves fine-grained details while capturing class-general features by updating only non-critical pixels.}
\label{fig:diff}
\end{figure}

To address this challenge, we propose a novel Distance-Based Representative (DBR) knowledge transfer technique that eliminates the need for traditional soft labels. DBR employs a distance-based approach to measure the discrepancy between predictions on synthetic data and one-hot encoded labels, simplifying the training process and reducing label storage requirements. For example, our method requires only 0.2 GB to store ImageNet1k (200 IPC), achieving a 500× reduction in storage while still delivering comparable results. By integrating DBR with NRR, our method enhances dataset distillation by capturing essential features while minimizing training complexity. This results in more compact and efficient datasets, well-suited for a variety of training environments. Figure \ref{fig:diff} illustrates the differences between our method and two popular large-scale dataset distillation techniques, RDED \cite{rded} and SRe$^2$L. It is evident that RDED focuses on instance-specific features without refinement, while SRe$^2$L updates all pixels to capture class-general features, often at the expense of fine-grained details. In contrast, our NRR-DD method effectively preserves fine-grained details by updating only non-critical pixels, while still capturing class-general features.

\textbf{Contributions.} Our major contributions are summarized as follows:
\begin{itemize}[noitemsep,nolistsep]
    \item We introduce the Non-Critical Region Refinement Dataset Distillation (NRR-DD) framework, which consists of three key stages: Critical-based Initial Data Discovery (CIDD), Non-Critical Region Refinement (CRR), and Relabeling. This approach generates synthetic data that captures both instance-specific fine-grained features and class-general patterns, significantly enhancing performance.
    \item We propose the Distance-Based Representative (DBR) method for knowledge transfer, eliminating the need for soft labels and drastically reducing memory requirements. Specifically, our method reduces storage requirements by \textit{500-fold} compared to soft labels on ImageNet1k, while recovering up to $80\%$ of the full performance (see Table \ref{tab:ab-cidd}).
    \item Experimental results demonstrate that our NRR method achieves state-of-the-art performance on both small- and large-scale datasets. Additionally, by storing only two distances per instance, it achieves comparable results across various settings.
\end{itemize}

\section{Related Works}

\noindent
\textbf{Dataset Distillation.} Several dataset distillation methods have been proposed recently \cite{ddgm, ddtm, idm, rded, tesla, sre2l}, which can be classified into two main categories. The first category, Class-General Feature-Based Methods, aims to capture class-wide features. For example, Gradient Matching \cite{ddgm, ddgm2} generates synthetic data by matching gradients across all samples in a class, while Distribution Matching \cite{idm, dddm2, dddm3} uses distribution prediction. Trajectory Matching \cite{ddtm, ddtm2} aligns training trajectories of original and synthetic data, and SRe$^2$L \cite{sre2l} recovers Batch Normalization statistics to capture class-general features. The second category, Instance-Specific Feature-Based Methods \cite{rded, mmdiff}, includes approaches such as RDED \cite{rded}, which extracts high-confidence patches and combines them to create synthetic data, and MDiff \cite{mmdiff}, which utilizes a diffusion model to generate fine-grained images tailored to the task. However, each approach has its limitations: methods focused on class-general features risk losing crucial instance-specific details, hindering fine-grained generalization, while instance-specific methods may neglect broader class patterns, leading to suboptimal class-level representation.

\noindent
\textbf{Large-Scale Dataset Distillation via Soft-Label Knowledge Transfer.} Recently, several methods have been proposed for large-scale datasets like ImageNet1k \cite{tesla, sre2l, rded}. However, all of these approaches require storing the soft labels of each augmented data point for training new student models, leading to significant memory storage overhead. For example, with ImageNet1k, storing the soft labels for 50 Images-Per-Class (IPC) requires approximately 30 GB, and for 200 IPC, over 120 GB of storage \cite{sre2l}. This highlights the need for novel knowledge transfer techniques to reduce memory storage requirements.

\section{Proposed Method}
\begin{figure*}[t]
\begin{center}
\includegraphics[width=\linewidth]{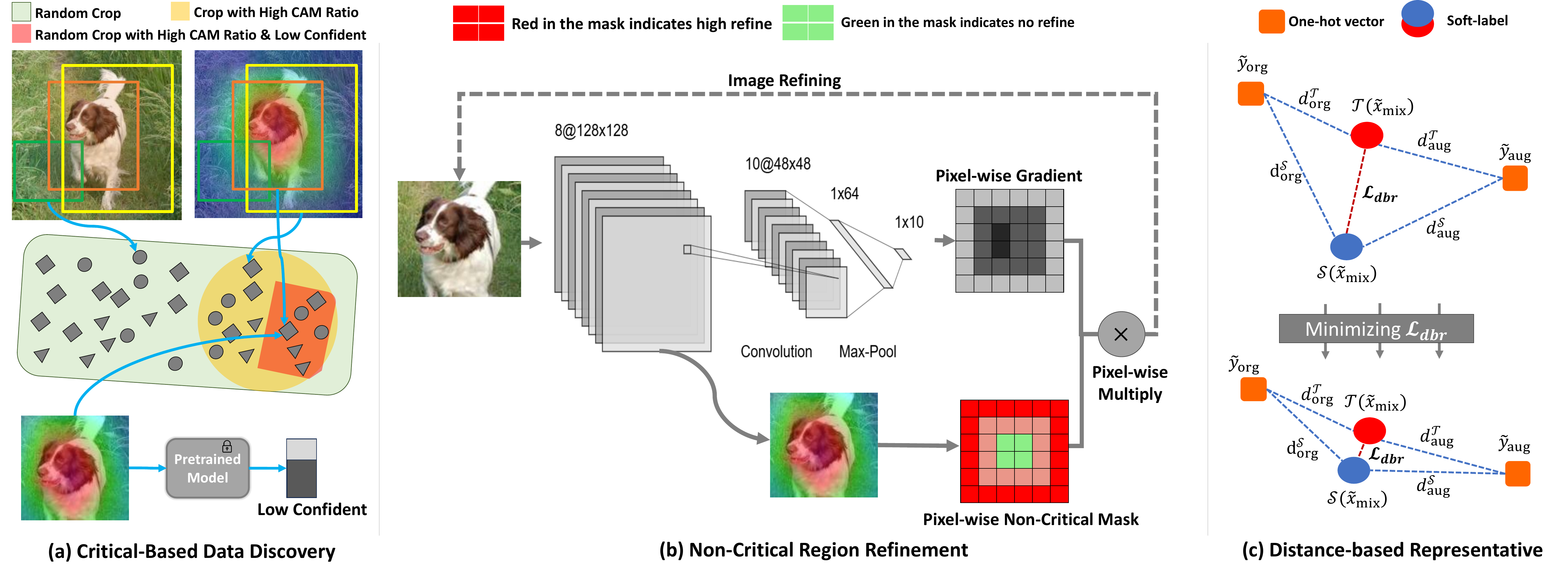}
\end{center}
\caption{The architecture of our NRR-DD consists of three key stages: (i) Critical-based Initial Data Discovery (Section \ref{sec:ids}), which selects patches with a high CAM ratio but low confidence level to capture instance-specific features; (ii) Non-Critical Region Refinement (Section \ref{sec:nrr}), where CAM \cite{cam} is used to identify and refine both critical and non-critical regions, preserving fine-grained details while enriching non-critical areas with class-general information; (iii) Knowledge Transfer, which aims to minimize the distance between $\mathcal{S}(\tilde{x}_\text{mix})$ (student prediction) and $\mathcal{T}(\tilde{x}_\text{mix})$ (pretrained teacher prediction or soft label) by reducing the distance between $d^T_\text{org}$ and $d^S_\text{org}$, as well as between $d^T_\text{aug}$ and $d^S_\text{aug}$. By storing only the two values, $d^T_\text{org}$ and $d^T_\text{aug}$, the new model can effectively mimic the performance of the pretrained one.}
\label{fig:nrrdd}
\end{figure*}

In this section, we first provide the necessary preliminaries for the Dataset Distillation Method (Section \ref{sec:pre}), followed by an introduction to our Non-Critical Region Refinement Dataset Distillation (NRR-DD) framework. The framework consists of three key stages: (i) \textbf{Critical-based Initial Data Discovery} (Section \ref{sec:ids}), which selects diverse, important patches to capture instance-specific features; (ii) \textbf{Non-Critical Region Refinement} (Section \ref{sec:nrr}), where Class Activation Mapping (CAM) \cite{cam, gradcam} is employed to identify and refine both critical and non-critical regions, preserving fine-grained details while enriching non-critical areas with class-general information; and (iii) \textbf{Knowledge Transfer via Relabeling} (Section \ref{sec:ktr}), in which synthetic images are relabeled for knowledge transfer to a student model. The overall architecture is shown in Figure \ref{fig:nrrdd}, and the pseudo code can be found in Algorithm \ref{alg:nrr-dd}.

\begin{algorithm}[t]
\caption{NRR-DD}
\label{alg:nrr-dd}
\kwInput{Pre-trained model $\mathcal{T}_{\theta_\mathcal{T}}$, training set $\mathcal{D} = \{(\vx_i, \vy_i)\}_{i=1}^m$}
\kwOutput{Synthetic dataset $\tilde{\mathcal{D}} = \{(\tilde{\vx}_i, \tilde{\vy}_i)\}_{i=1}^n$}
Initial $\tilde{\mathcal{D}} = \emptyset$\;
\Comment{Critical-based Initial Data Discovery}
\ForEach{$(\vx, \vy) \sim \mathcal{D}$}{
    Calculate CAMs matrix $C(\vx)$ for $\vx$\;
    Crop $\vx$ into $k$ patches $\{p_1, \dots, p_k\}$\;
    Select the top $t$ patches with the highest values in the CAMs matrix\;
    Resize and store them in the patch pool $\mathcal{P}$\;
}
\ForEach{$p \sim \mathcal{D}$}{
    Compute the score $s(p) = \mathcal{L}_{ce}(p)$\;
}
Select the top $g = \beta \times IPC$ patches with the lowest scores\;
\For{$i = 1$ to $IPC$}{
    Combine $\beta$ patches into $\tilde{\vx}$, store ($\tilde{\vx}, \tilde{\vy}$) in $\tilde{\mathcal{D}}$
}
\Comment{Non-Critical Region Refinement}
\ForEach{$(\tilde{\vx}_\text{org}, \tilde{\vy}_\text{org}) \sim \tilde{\mathcal{D}}$}{
    \For{$I$ iterations}{
        $(\tilde{\vx}_\text{aug}, \tilde{\vy}_\text{aug}) \sim \tilde{\mathcal{D}}$\;
        $\tilde{\vx}_\text{mix} = A(\tilde{\vx}_\text{org}, \tilde{\vx}_\text{aug})$\;
        Update $\tilde{\vx}_\text{org}$ by Eq. \ref{eq:lcnew2}\;
    }
    $\tilde{\vy}_\text{soft} = \mathcal{T}(\tilde{\vx}_\text{mix})$\;
    $d^T_\text{org} = \mathcal{L}_{ce}(\tilde{\vy}_\text{soft}, \tilde{\vy}_\text{org})$\;
    $d^T_\text{aug} = \mathcal{L}_{ce}(\tilde{\vy}_\text{soft}, \tilde{\vy}_\text{aug})$\;
    Store ($\tilde{\vx}_\text{org}, (\tilde{\vy}_\text{org}, \tilde{\vy}_\text{aug}), (d^T_\text{org}, d^T_\text{aug})$) in $\tilde{\mathcal{D}}$\;
}
\Comment{Distance-Based Representative Knowledge Transfer}
\ForEach{$(\tilde{\vx}_\text{org}, (\tilde{\vy}_\text{org}, \tilde{\vy}_\text{aug}), (d^T_\text{org}, d^T_\text{aug})) \sim \tilde{\mathcal{D}}$}{
    Update new model $\mathcal{S}$ by Eq. \ref{eq:ls}\;
}
\end{algorithm}

\subsection{Preliminaries}
\label{sec:pre}
Consider a training dataset $\mathcal{D} = \{(\vx_i, \vy_i)\}_{i=1}^m$, where each input $\vx_i \in \mathbb{R}^{c \times h \times w}$ represents a sample, and $\vy_i \in \{1, \dots, K\}$ denotes its label. Let $\mathcal{T} = \mathcal{T}_{\theta_\mathcal{T}}$ be a pretrained model on $\mathcal{D}$. The goal of dataset distillation is to generate a synthetic dataset $\tilde{\mathcal{D}} = \{(\tilde{\vx}_i, \tilde{\vy}_i)\}_{i=1}^n$ (with $n \ll m$) that retains the essential information from $\mathcal{D}$, enabling a new model $\mathcal{S}$ to achieve performance comparable to that of $\mathcal{T}$.

\subsection{Critical-based Initial Data Discovery}
\label{sec:ids}

In this section, instead of using Gaussian noise to generate synthetic data for $\mathcal{M}$, we propose training-free techniques to more effectively select initial data. The motivation behind this approach is that real data inherently contains fine-grained, instance-specific features, which are crucial for training on large-scale datasets \cite{rded,mmdiff,shapeformer,ppsn,pisd,csax}.

Given the training dataset $\mathcal{D} = \{(\vx_i, \vy_i)\}_{i=1}^m$, for each data example $(\vx, \vy) \in \mathcal{D}$, we first compute the class activation mapping (CAM) \cite{cam}, producing a matrix $C(\vx, \vy)$ for the image $\vx$ and class $\vy$:
\begin{equation} 
C(\vx, \vy) = \sum_k \vw_k^{\vy} \mathcal{T}_k(\vx, \vy), 
\label{eq:cam}
\end{equation}
where $\vw_k^{\vy}$ represents the $k^\text{th}$ weight in the final classification head for class $\vy$, and $\mathcal{T}_k$ denotes the $k^\text{th}$ feature map from the final layers of the model.

Next, the images $\vx$ are randomly cropped to extract multiple patches, and the top $t$ patches, which contain the highest values in the class activation map, are selected. These patches are then resized to the full image size and stored in the patch pool $\mathcal{P}$. Since the selection is based on the highest values in the class activation map, the chosen patches capture the most important information from the original images.

For each patch $\vp$ in $\mathcal{P}$, we feed it into $\mathcal{T}$ to obtain a confidence score $s = \mathcal{T}(\vp)$, which represents the highest prediction probability. Unlike RDED \cite{rded}, which selects patches with the highest scores, we choose the top $g = \beta \times IPC$ patches with the lowest scores, where $\beta = 1, 4, 9, \dots$ denotes the number of patches used to form a single synthetic image, and $IPC$ specifies the number of images per class. Notably, our strategy of selecting the lowest-scoring patches identifies the hardest-to-learn samples, while the method of selecting patches with the highest values in the class activation map ensures that the patches contain important information. This provides more opportunity and flexibility in the next phase to further refine the chosen synthetic images. Figure \ref{fig:diff} visualizes the images selected by both the highest- and lowest-scoring strategies.

Finally, similar to RDED \cite{rded}, we construct each synthetic image by combining $\beta$ patches. The selected patches are resized to $1/\beta$ of their original size and then combined to create $IPC$ synthetic images per class, all of which are stored in the dataset $\tilde{\mathcal{D}}$. Unlike RDED \cite{rded}, our synthetic images in $\tilde{\mathcal{D}}$ are refined to include both \textit{fine-grained, instance-specific features} and \textit{class-general features}.

\subsection{Non-Critical Region Refinement}
\label{sec:nrr}
In contrast to RDED \cite{rded}, our synthetic images in $\tilde{\mathcal{D}}$ are refined to incorporate both \textit{detailed, instance-specific} features and \textit{broader, class-general} features. Specifically, for each $(\tilde{\vx}, \tilde{\vy}) \in \tilde{\mathcal{D}}$, the image $\tilde{\vx}$ is refined according to the following loss function:
\begin{align}
    \mathcal{L}_{C} &= \mathcal{L}_{ce}(\mathcal{T}(\tilde{\vx}), \tilde{\vy}) + \alpha_{bn} \mathcal{L}_{bn}(\mathcal{T}(\tilde{\vx})), 
    \label{eq:lc}
\end{align}
where $\alpha_{bn}$ is a parameter.

In this framework, $\mathcal{L}_{ce}$ represents the cross-entropy (CE) loss, which aims to move $\tilde{\vx}$ with the smallest confidence score into the teacher’s high-confidence regions. The batch normalization regularization ($\mathcal{L}_{bn}$) \cite{nayer,sre2l,muse}, a standard DFKD loss, aligns the mean and variance at the \texttt{BatchNorm} layer with its running mean and variance:
\begin{equation}
    \mathcal{L}_{bn} = \sum\limits_{l} \left( \|\mu_l(\tilde{\vx}) - \mu_l\| + \|\sigma^2_l(\tilde{\vx}) - \sigma^2_l\| \right),
\end{equation}
where $\mu_l(\tilde{\vx})$ and $\sigma^2_l(\tilde{\vx})$ are the mean and variance of the $l$-th \texttt{BatchNorm} layer of $\mathcal{T}$, and $\mu_l$ and $\sigma^2_l$ are the running mean and variance of the $l$-th \texttt{BatchNorm} layer in $\mathcal{T}$.

\begin{figure*}[t]
\begin{center}
\includegraphics[width=\linewidth]{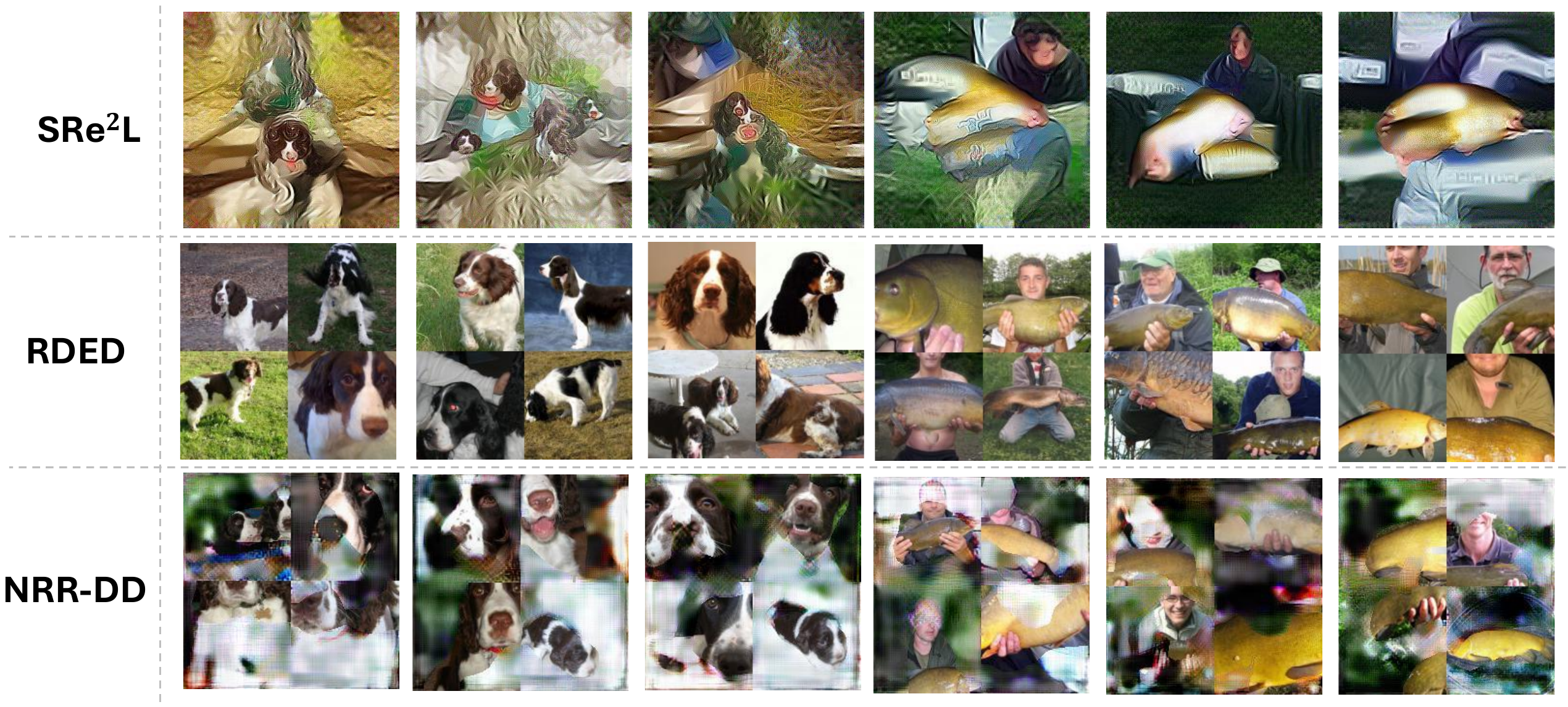}
\end{center}
\caption{Visualization of images from the `tench' and `English springer' classes synthesized using various dataset distillation methods, including SRe$^2$L \cite{sre2l}, RDED \cite{rded}, and our NRR-DD. For additional visualizations, please refer to the \textbf{Supplementary Material}.}
\label{fig:visual}
\end{figure*}

\noindent
\textbf{Non-Critical Region.} A naive approach to refining $\tilde{\vx}$ would involve optimizing all pixels in $\tilde{\vx}$ using the gradient of $\mathcal{L}_C$. However, this would lead to significant changes in the images, resulting in the loss of fine-grained, instance-specific features, as shown in Figure \ref{fig:diff}. Instead, we propose the \textit{Non-Critical Region Refinement Dataset Distillation} (NRR-DD) method, which preserves instance-specific and fine-grained regions in the synthetic data while enriching non-critical regions with more class-general information. This approach enables our models to utilize all pixel information to capture both types of features, thereby enhancing overall performance.

Given a synthetic image $\tilde{\vx}$ with label $\tilde{\vy}$, we use CAM to create a \textit{non-critical mask} $M$ of the same size as $\tilde{\vx}$. This mask assigns low or zero values to high-importance pixels in $\tilde{\vx}$ and higher values to less important pixels. The objective is to control pixel updates in $\tilde{\vx}$ by limiting updates to important pixels in order to preserve instance-specific features, while allowing less important pixels to update more significantly, thus enhancing the learning of class-general features. The process is detailed as follows.

We first generate the CAM matrix $C$ of $(\tilde{\vx}, \tilde{\vy})$ using Eq.~\ref{eq:cam}. Subsequently, we calculate the pixel-wise non-critical mask $M$ matrix using the following formula:
\begin{equation}
    M = \max\{0, \epsilon - C\}.
\end{equation}
Here, $\epsilon$ serves as the upper threshold for the method. For any value $c$ in $C$ that exceeds $\epsilon$, its corresponding value in $M$ will be set to 0; otherwise, the value will be $\epsilon - c$. Since higher class activation values correspond to lower non-critical scores, the matrix $M$ will be used to weight the gradient update for each pixel in $\tilde{\vx}$. With each gradient update, the image $\tilde{\vx}$ is updated as follows:
\begin{equation}
    \tilde{\vx} = \tilde{\vx} - M \times \eta \nabla_{\tilde{\vx}} \mathcal{L}_{\mathcal{C}},
\end{equation}
where $ \nabla_{\tilde{\vx}} \mathcal{L}_{\mathcal{C}}$ represents the gradient of the loss function $\mathcal{L}_{\mathcal{C}}$ with respect to the image $\tilde{\vx}$.

\subsection{Knowledge Transfer via Relabeling}
\label{sec:ktr}

\noindent
\noindent
\textbf{Soft-Label Knowledge Transfer.} Previous methods \cite{sre2l, tesla, rded} store soft labels generated by a pretrained teacher model $\mathcal{T}$ to train a student model $\mathcal{S}$ by minimizing the Kullback-Leibler (KL) divergence between the student model's predictions and the teacher's soft labels. These soft labels are generated through augmentation techniques such as CutMix \cite{cutmix} and Mixup \cite{mixup}. 

Given a pair of original image $(\tilde{\vx}_\text{org}, \tilde{\vy}_\text{org})$ and augmented image $(\tilde{\vx}_\text{aug}, \tilde{\vy}_\text{aug})$ in the synthetic dataset $\tilde{\mathcal{D}}$, we apply CutMix or Mixup to create the mixed image and compute its soft label as:  
\begin{align}
    \tilde{\vx}_\text{mix} &= A(\tilde{\vx}_\text{org}, \tilde{\vx}_\text{aug}) \\
    \tilde{\vy}_\text{soft} &= \mathcal{T}(\tilde{\vx}_\text{mix}).
\end{align}
Here, $A$ represents an augmentation method such as CutMix \cite{cutmix} or Mixup \cite{mixup}, and $\tilde{\vy}_{\text{soft}}$ denotes the soft labels produced by the mixed image $\tilde{\vx}_{\text{mix}}$. 

Therefore, for each pair of original image $(\tilde{\vx}_\text{org}, \tilde{\vy}_\text{org})$ and augmented image $(\tilde{\vx}_\text{aug}, \tilde{\vy}_\text{aug})$, we need to store the indices of the two images in the synthetic dataset, the specification of the mixing method $A$ used to construct the mixed image $\tilde{\vx}_\text{mix}$, and the soft label $\tilde{\vy}_\text{soft}$. However, this approach requires substantial memory to store the soft labels, especially when large datasets and numerous augmentations are involved.
\begin{table*}[t]
\centering
\begin{adjustbox}{width=\linewidth}
\begin{tabular}{@{}ll|ccccc|ccc|ccc@{}}
\toprule
              &     & \multicolumn{5}{c}{ConvNet}                                                                                                                       & \multicolumn{3}{c}{Resnet18}                                                    & \multicolumn{3}{c}{ResNet-101}                                                  \\ \midrule
              & IPC & MTT                            & IDM                            & TESLA      & RDED       & NRR-DD                & SRe$^2$L      & RDED       & NRR-DD                & SRe$^2$L      & RDED       & NRR-DD                \\ \midrule
CIFAR10       & 1   & \multicolumn{1}{l}{46.3 ± 0.8} & \multicolumn{1}{l}{45.6 ± 0.7} & 48.5 ± 0.8                & 23.5 ± 0.3               & \textbf{48.4 ± 0.4}      & 16.6 ± 0.9                & 22.9 ± 0.4               & \textbf{30.3 ± 0.4}      & 13.7 ± 0.2                & 18.7 ± 0.1               & \textbf{25.7 ± 0.3}      \\
              & 10  & \multicolumn{1}{l}{65.3 ± 0.7} & \multicolumn{1}{l}{65.3 ± 0.7} & 66.4 ± 0.8                & 50.2 ± 0.3               & \textbf{66.7 ± 0.4}      & 29.3 ± 0.5                & 37.1 ± 0.3               & \textbf{72.2 ± 0.4}      & 24.3 ± 0.6                & 33.7 ± 0.3               & \textbf{65.1 ± 0.3}      \\
              & 50  & 71.6 ± 0.2                     & 67.5 ± 0.1                     & 72.6 ± 0.7                & 68.4 ± 0.1               & \textbf{73.1 ± 0.1}      & 45.0 ± 0.7                & 62.1 ± 0.1               & \textbf{84.1 ± 0.1}      & 34.9 ± 0.1                & 51.6 ± 0.4               & \textbf{78.2 ± 0.4}      \\ \midrule
CIFAR100      & 1   & 24.3 ± 0.3                     & 20.1 ± 0.3                     & 24.8 ± 0.5                & 19.6 ± 0.3               & \textbf{27.3 ± 0.3}      & 6.6 ± 0.2                 & 11.0 ± 0.3               & \textbf{33.3 ± 0.3}      & 6.2 ± 0.0                 & 10.8 ± 0.1               & \textbf{32.9 ± 0.3}      \\
              & 10  & 40.1 ± 0.4                     & 45.1 ± 0.1                     & 41.7 ± 0.3                & 48.1 ± 0.3               & \textbf{55.7 ± 0.2}      & 27.0 ± 0.4                & 42.6 ± 0.2               & \textbf{62.7 ± 0.2}      & 30.7 ± 0.3                & 41.1 ± 0.2               & \textbf{58.3 ± 0.2}      \\
              & 50  & 47.7 ± 0.2                     & 50.0 ± 0.2                     & 47.9 ± 0.3                & 57.0 ± 0.1               & \textbf{61.1 ± 0.1}      & 50.2 ± 0.4                & 62.6 ± 0.1               & \textbf{67.1 ± 0.1}      & 56.9 ± 0.1                & 63.4 ± 0.3               & \textbf{65.1 ± 0.3}      \\  \midrule
Tiny-ImageNet & 1   & 8.8 ± 0.3                      & 10.1 ± 0.2                     & -                         & 12.0 ± 0.1               & \textbf{20.4 ± 0.2}      & 2.62 ± 0.1                & 9.7 ± 0.4                & \textbf{13.5 ± 0.2}      & 1.9 ± 0.1                 & 3.8 ± 0.1                & \textbf{10.1 ± 0.1}      \\
              & 10  & 23.2 ± 0.2                     & 21.9 ± 0.3                     & -                         & 39.6 ± 0.1               & \textbf{44.3 ± 0.2}      & 16.1 ± 0.2                & 41.9 ± 0.2               & \textbf{45.2 ± 0.2}      & 14.6 ± 1.1                & 22.9 ± 3.3               & \textbf{26.1 ± 3.3}      \\
              & 50  & 28.0 ± 0.3                     & 27.7 ± 0.3                     & -                         & 47.6 ± 0.2               & \textbf{50.2 ± 0.1}      & 41.1 ± 0.4                & 58.2 ± 0.1               & \textbf{61.2 ± 0.1}      & 42.5 ± 0.2                & 41.2 ± 0.4               & \textbf{46.2 ± 0.4}      \\ \midrule
ImageNette    & 1   & 47.7 ± 0.9                     &     -                           & -                         & 33.8 ± 0.8               & \textbf{39.3 ± 0.9}      & 19.1 ± 1.1                & 35.8 ± 1.0               & \textbf{40.1 ± 0.9}      & 15.8 ± 0.6                & 25.1 ± 2.7               & \textbf{28.1 ± 2.7}      \\
              & 10  & 63.0 ± 1.3                     &           -                     & -                         & 63.2 ± 0.7               & \textbf{68.3 ± 0.6}      & 29.4 ± 3.0                & 61.4 ± 0.4               & \textbf{66.2 ± 0.6}      & 23.4 ± 0.8                & 54.0 ± 0.4               & \textbf{56.0 ± 0.4}      \\
              & 50  &                    -            &           -                     & -                         & 83.8 ± 0.2               & \textbf{86.5 ± 0.3}      & 40.9 ± 0.3                & 80.4 ± 0.4               & \textbf{85.6 ± 0.3}      & 36.5 ± 0.7                & 75.0 ± 1.2               & \textbf{78.0 ± 1.2}      \\ \midrule
ImageNet1k    & 1   &             -                   &   -                             & 7.7 ± 0.2                 & 6.4 ± 0.1                & \textbf{11.2 ± 0.2}      & 0.1 ± 0.1                 & 6.6 ± 0.2                & \textbf{11.6 ± 0.2}      & 0.6 ± 0.1                 & 5.9 ± 0.4                & \textbf{12.2 ± 0.4}      \\
              & 10  &            -                    &             -                   & 17.8 ± 1.3                & 20.4 ± 0.1               & \textbf{25.6 ± 0.2}      & 21.3 ± 0.6                & 42.0 ± 0.1               & \textbf{46.1 ± 0.2}      & 30.9 ± 0.1                & 48.3 ± 1.0               & \textbf{51.3 ± 1.0}      \\
              & 50  &          -                      &             -                   & 27.9 ± 1.2                & 38.4 ± 0.2               & \textbf{42.1 ± 0.1}      & 46.8 ± 0.2                & 56.5 ± 0.1               & \textbf{60.2 ± 0.1}      & 60.8 ± 0.5                & 61.2 ± 0.4               & \textbf{64.3 ± 0.4}      \\ \bottomrule
\end{tabular}
\end{adjustbox}
\caption{Comparison with state-of-the-art (SOTA) dataset distillation baselines. Identical neural networks are used for both dataset distillation and evaluation. Following \cite{rded,sre2l}, ConvNets used for distillation are Conv-3 for CIFAR10 and CIFAR100, Conv-4 for Tiny-ImageNet and ImageNet-1K, Conv-5 for ImageNette and ImageWoof, and Conv-6 for ImageNet-100. MTT and TESLA use down-sampled images for distillation to 224 $\times$ 224 images. SRe2L and RDED use ResNet-18 for distillation and retrieval, and are evaluated on ResNet-18 and ResNet-101. Entries marked with “-” indicate scalability issues. See \textbf{Supplementary Material} for further details.}
\label{tab:full}
\end{table*}

\noindent
\textbf{Distance-Based Representative Knowledge Transfer.} To address the memory limitations of storing soft labels, we propose a more memory-efficient approach. Instead of storing the soft labels $\tilde{\vy}_{\text{soft}}$, each of which consists of $1,000$ real numbers for ImageNet1K, we store only two real numbers representing the cross-entropy (CE) divergences between the soft label and the one-hot vectors of $\tilde{\vy}_{\text{aug}}$ and $\tilde{\vy}_{\text{org}}$. 

Specifically, for each pair of an original image $(\tilde{\vx}_{\text{org}}, \tilde{\vy}_{\text{org}})$ and an augmented image $(\tilde{\vx}_{\text{aug}}, \tilde{\vy}_{\text{aug}})$ in the synthetic dataset $\tilde{\mathcal{D}}$, we apply CutMix or Mixup to create the mixed image $\tilde{\vx}_{\text{mix}}$ and compute its soft label $\tilde{\vy}_{\text{soft}}$ using the teacher model $\mathcal{T}$. We then calculate and store the cross-entropy (CE) divergences between the soft label and the one-hot vectors of $\tilde{\vy}_{\text{aug}}$ and $\tilde{\vy}_{\text{org}}$ as follows:
\begin{align}
    d^T_\text{org} &= \mathcal{L}_{ce}(\tilde{\vy}_\text{soft}, \tilde{\vy}_\text{org}), \nonumber \\
    d^T_\text{aug} &= \mathcal{L}_{ce}(\tilde{\vy}_\text{soft}, \tilde{\vy}_\text{aug}).
\end{align}
Moreover, for each pair of original and augmented images, we store their indices in the synthetic dataset, the details of the data mixing, and the two divergences $d^T_{\text{org}}$ and $d^T_{\text{aug}}$ (see Figure \ref{fig:nrrdd}(c) for visualization). Subsequently, we train the student model $\mathcal{S}$ by minimizing:
\begin{align}
    \mathcal{L}_{\mathcal{S}} &= \mathcal{L}_{sce} + \alpha_{dbr}\mathcal{L}_{dbr},\label{eq:ls} \\
    \mathcal{L}_{sce} &= max\{0,d^S_\text{org} - r\} + max\{0,d^S_\text{aug} - r\}, \nonumber\\
    \mathcal{L}_{dbr} &= |d^S_\text{org} - d^T_\text{org}| + |d^S_\text{aug} - d^T_\text{aug}|, \nonumber 
\end{align}
where $r$ is a threshold and $d^S$ is calculated as follows:
\begin{align}
    d^S_\text{org} &= \mathcal{L}_{ce}(\mathcal{S}(\tilde{\vx}_\text{mix}),\tilde{\vy}_\text{org}) \nonumber\\
d^S_\text{aug} &= \mathcal{L}_{ce}(\mathcal{S}(\tilde{\vx}_\text{mix}),\tilde{\vy}_\text{aug}).
\end{align}

In Eq. \ref{eq:ls}, $\mathcal{L}_{\text{sce}}$ denotes the soft cross-entropy loss function, which encourages the student model to predict the mixed instance $\tilde{\vx}_{\text{mix}}$ as a blend of the two labels, $\tilde{\vy}_{\text{org}}$ and $\tilde{\vy}_{\text{aug}}$. This helps prevent excessive confidence by imposing a threshold $r$, which is especially crucial when using techniques such as CutMix or MixUp, where the soft label often blends multiple classes. This mechanism reduces the risk of the model becoming overly confident in any single class. Additionally, $\mathcal{L}_{\text{dbr}}$ represents the distance-based representative loss, which allows the student model $\mathcal{S}$ to replicate the teacher model’s divergences from its predictions on the mixed image $\tilde{\vx}_{\text{mix}}$ to $\tilde{\vy}_{\text{org}}$ and $\tilde{\vy}_{\text{aug}}$. This loss ensures consistency between the teacher and student models, both in terms of learned feature representations and decision boundaries.

\noindent
\textbf{Memory Reduction.} It is important to note that, unlike previous models that require storing \textbf{augmentation information} to generate $\tilde{\vx}_\text{mix}$, along with the indices and labels for both $\tilde{\vx}_\text{org}$ and $\tilde{\vx}_\text{aug}$, our approach only necessitates storing two additional distances, $d^T_\text{org}$ and $d^T_\text{aug}$, for the loss function. This significantly reduces memory requirements.

\noindent
\textbf{Label Refinement.} To further enhance the benefits of Distance-Based Representative (DBR), we propose refining the images with an additional term. Instead of training $\tilde{\vx}_\text{org}$ using the formula from Eq. \ref{eq:lc}, we train $\tilde{\vx}_\text{org}$ with the following formulation:
\begin{align}
    \mathcal{L}_C &= \mathcal{L}_\text{org} + \alpha_{lr} \mathcal{L}_{lr}, 
    \label{eq:lcnew2}\\
    \mathcal{L}_\text{org} &= \mathcal{L}_{ce}(\mathcal{T}(\tilde{\vx}_\text{org}), \tilde{\vy}_\text{org}) + \alpha_{bn}\mathcal{L}_{bn}(\mathcal{T}(\tilde{\vx}_\text{org})) \nonumber \\
    \mathcal{L}_{lr} &= \max\{0, d^T_\text{org} - r\} + \max\{0, d^T_\text{aug} - r\}  
    \label{eq:lcnew}
\end{align}
Minimizing $\mathcal{L}_{lr}$ ensures that the prediction of $\mathcal{T}$ on $\tilde{\vx}_\text{mix}$, which will be used for training the student model, is focused solely on the two designated classes, facilitating alignment with the DBR function.

In Figure \ref{fig:visual}, we visualize the images generated by our NRR-DD method alongside several state-of-the-art approaches. It is clear that our method captures both instance-specific and class-general features, leading to improved performance.

\section{Experiment}

\begin{table*}[t]
\centering
\begin{adjustbox}{width=\linewidth}
\begin{tabular}{@{}l|l|l|cc|c|cccc@{}}
\toprule
                            &                           &     & \multicolumn{2}{c|}{Using Soft Label} & Using One-Hot & \multicolumn{3}{c}{Using Compact Label}     &                    \\ \midrule
Dataset                     & Architecture              & IPC & RDED (SL)         & NRR-DD(SL)       & RDED (OH)     & RDED (CL)  & NRR-DD (DBR) & NRR-DD (DBR+LR) & Recover \\ \midrule
\multirow{6}{*}{ImageNette} & \multirow{3}{*}{ConvNet}  & 1   & 33.8 ± 0.8   & \textbf{39.3 ± 0.9}   & 16.3 ± 0.3    & 23.2 ± 0.5 & {\ul 32.3 ± 0.4} & \textbf{34.5 ± 0.5} & 79\%               \\
                            &                           & 10  & 63.2 ± 0.7   & \textbf{68.3 ± 0.6}   & 27.3 ± 0.2    & 37.3 ± 0.4 & {\ul 52.2 ± 0.4} & \textbf{55.1 ± 0.4} & 68\%               \\
                            &                           & 50  & 83.8 ± 0.2   & \textbf{86.5 ± 0.3}   & 41.2 ± 0.3    & 54.6 ± 0.3 & {\ul 66.9 ± 0.2} & \textbf{69.2 ± 0.2} & 62\%               \\  \cmidrule(l){2-10} 
                            & \multirow{3}{*}{Resnet18} & 1   & 35.8 ± 1.0   & \textbf{40.1 ± 0.9}   & 16.2 ± 0.4    & 22.2 ± 0.4 & {\ul 34.6 ± 0.6} & \textbf{36.2 ± 0.4} & 84\%               \\
                            &                           & 10  & 61.4 ± 0.4   & \textbf{66.2 ± 0.6}   & 25.4 ± 0.3    & 34.1 ± 0.4 & {\ul 54.7 ± 0.4} & \textbf{57.1 ± 0.4} & 78\%               \\
                            &                           & 50  & 80.4 ± 0.4   & \textbf{85.6 ± 0.3}   & 41.3 ± 0.3    & 52.1 ± 0.2 & {\ul 69.4 ± 0.2} & \textbf{72.2 ± 0.3} & 70\%               \\  \midrule
\multirow{6}{*}{ImageNet1k} & \multirow{3}{*}{ConvNet}  & 1   & 6.4 ± 0.1    & \textbf{11.2 ± 0.2}   & 2.4 ± 0.5     & 3.1 ± 0.5  & {\ul 6.2 ± 0.5}  & \textbf{8.5 ± 0.5}  & 69\%               \\
                            &                           & 10  & 20.4 ± 0.1   & \textbf{25.6 ± 0.2}   & 8.3 ± 0.4     & 12.5 ± 0.3 & {\ul 16.1 ± 0.3} & \textbf{19.2 ± 0.5} & 63\%               \\
                            &                           & 50  & 38.4 ± 0.2   & \textbf{42.1 ± 0.1}   & 14.1 ± 0.3    & 22.3 ± 0.3 & {\ul 26.1 ± 0.1} & \textbf{28.1 ± 0.3} & 50\%               \\  \cmidrule(l){2-10} 
                            & \multirow{3}{*}{Resnet18} & 1   & 6.6 ± 0.2    & \textbf{11.6 ± 0.2}   & 2.3 ± 0.5     & 3.2 ± 0.5  & {\ul 7.5 ± 0.3}  & \textbf{8.9 ± 0.5}  & 71\%               \\
                            &                           & 10  & 42.0 ± 0.1   & \textbf{46.1 ± 0.2}   & 16.3 ± 0.7    & 22.2 ± 0.4 & {\ul 34.3 ± 0.5} & \textbf{37.2 ± 0.4} & 70\%               \\
                            &                           & 50  & 56.5 ± 0.1   & \textbf{60.2 ± 0.1}   & 32.4 ± 0.5    & 39.3 ± 0.2 & {\ul 45.1 ± 0.3} & \textbf{49.2 ± 0.5} & 60\%               \\\bottomrule 
\end{tabular}

\end{adjustbox}
\caption{Comparison of various relabeling methods on large-scale datasets, including ImageNette and ImageNet1k. Bold values indicate the highest scores, while underlined values indicate the second-highest scores. `SL' denotes the use of soft labels, `OH' represents one-hot vector labels, and `CL' signifies compact labels (with 2 classes for fair comparison to our DBR). `DBR' refers to our method using distance-based representation without label refinement, and `DBR+LR' indicates our method with distance-based representation combined with label refinement. The Recover rate is calculate by using (DBR - One-hot)/(Soft-label - One-hot)}
\label{tab:dr}
\end{table*}

This section evaluates the effectiveness of our proposed method compared to state-of-the-art techniques across various datasets and neural architectures, accompanied by comprehensive ablation studies.

\subsection{Experimental Setting}
For large-scale datasets, we assessed our method using two popular pairs of backbones: ResNet34/ResNet18 \cite{resnet} and ResNet50/MobileNetV2 \cite{mbnetv2}, applied to three well-known benchmarks: Tiny-ImageNet \cite{tin}, which consists of 200 object categories, with 500 training images, 50 validation images, and 50 test images per category, all resized to $64 \times 64$ pixels; ImageNet1k \cite{imagenet}, containing 1,000 object categories and over 1.2 million labeled training images, along with its subset Imagenette, which includes 10 sub-classes. For small-scale datasets, we ran experiments with ResNet \cite{resnet}, VGG \cite{vgg}, and WideResNet (WRN) \cite{wrn} across CIFAR-10 and CIFAR-100 \cite{c10}. Both CIFAR-10 and CIFAR-100 consist of 60,000 images (50,000 for training and 10,000 for testing), with 10 and 100 categories, respectively, and all images have a resolution of $32 \times 32$ pixels. Consistent with previous research, we set the IPC to 1, 10, and 50. All experiments were run on a single NVIDIA A100 40 GB GPU. The details of the model architectures, parameters, and additional experimental results are provided in the \textbf{Supplementary Material}.


\noindent
\textbf{Compared Baselines.} We focus on comparing our method to state-of-the-art dataset distillation methods:
\begin{itemize}[noitemsep, nolistsep]
    \item MTT \cite{ddtm}, the first to propose trajectory matching;
    \item IDM \cite{idm}, the first to introduce distribution matching;
    \item TESLA \cite{tesla}, the first method to scale up to full ImageNet1k;
    \item SRe$^2$L \cite{sre2l}, a method that efficiently scales to ImageNet-1k and significantly outperforms existing approaches on large, high-resolution datasets;
    \item RDED \cite{rded}, a recent paper that uses instance-specific features with fine-grained details to improve large-scale dataset distillation, which we consider the closest baseline.
\end{itemize}

\subsection{Main Result}

\begin{table}[hpt]
\centering
\begin{adjustbox}{width=\linewidth}
\begin{tabular}{@{}lc|ccc@{}}
\toprule
\multicolumn{2}{c}{Teacher\textbackslash  Student} & Resnet18            & MobileNetV2         & EfficientNet-B0     \\ \midrule
Resnet18              & SRe$^2$L       & 21.7 ± 0.6          & 15.4 ± 0.2          & 11.7 ± 0.2          \\
                      & RDED        & 42.3 ± 0.6          & 40.4 ± 0.1          & 31.0 ± 0.1          \\
                      & NRR-DD      & \textbf{46.1 ± 0.2} & \textbf{45.0 ± 0.2} & \textbf{34.2 ± 0.1} \\\midrule
MobileNetV2           & SRe$^2$L       & 19.7 ± 0.1          & 10.2 ± 2.6          & 9.8 ± 0.4           \\
                      & RDED        & 34.4 ± 0.2          & 33.8 ± 0.6          & 24.1 ± 0.8          \\
                      & NRR-DD      & \textbf{36.2 ± 0.2} & \textbf{37.2 ± 0.1} & \textbf{27.3 ± 0.7} \\\midrule
EfficientNet-B0       & SRe$^2$L       & 25.2 ± 0.2          & 20.5 ± 0.2          & 11.4 ± 2.5          \\
                      & RDED        & 42.8 ± 0.5          & 43.6 ± 0.2          & 33.3 ± 0.9          \\
                      & NRR-DD      & \textbf{47.2 ± 0.2} & \textbf{45.6 ± 0.3} & \textbf{35.1 ± 0.2} \\ \bottomrule
\end{tabular}
\end{adjustbox}
\caption{Evaluation of ImageNet-1K top-1 accuracy for cross-architecture generalization. Datasets are distilled using ResNet-18, EfficientNet-B0, and MobileNet-V2, and transferred across different architectures. Note that experiments for SRe$^2$L could not be conducted when the distillation model lacks batch normalization \cite{sre2l}. All methods are evaluated with IPC = 10.}
\label{tab:cross}
\end{table}

\noindent
\textbf{Large-scale Dataset.} To demonstrate the effectiveness of our methods in real-world applications, we first compared them with various baselines on large-scale datasets such as ImageNet1k, Tiny ImageNet, and ImageNette. The results in Table \ref{tab:full} show that our methods outperform all the compared baselines in every scenario. For instance, our method achieves 60.2\% accuracy on ImageNet1k using ResNet18, which is 4\% higher than RDED and 14\% higher than SRe$^2$L. These improvements highlight the superiority of our approach in handling large-scale datasets and its ability to significantly outperform existing methods.

\noindent
\textbf{Small-scale Dataset.} To validate the robustness of our NRR-DD, we compared it with various baselines on small-scale datasets, including CIFAR-10 and CIFAR-100. As shown in Table \ref{tab:full}, our NRR-DD demonstrates significant improvements over the baselines. For example, with 10 IPC, our method achieves 72.2\% accuracy on CIFAR-10 using ResNet18, which is 35\% higher than the current state-of-the-art, RDED, which achieves 37.1\%. Similarly, on CIFAR-100, our method shows a 20\% improvement with 10 IPC. These results highlight the effectiveness and generalization ability of our method on small-scale datasets.

\begin{table}[h]
\centering
\begin{adjustbox}{width=1\linewidth}
\begin{tabular}{@{}l|cccccc@{}}
\toprule
        & \multicolumn{3}{c}{ConvNet}                                                                 & \multicolumn{3}{c}{Resnet18}                                                                \\ \midrule
        & \multicolumn{1}{l}{CIFAR10} & \multicolumn{1}{l}{CIFAR100} & \multicolumn{1}{l}{ImageNet1k} & \multicolumn{1}{l}{CIFAR10} & \multicolumn{1}{l}{CIFAR100} & \multicolumn{1}{l}{ImageNet1k} \\
RDED (baseline)    & 50.2 ± 0.3                  & 48.1 ± 0.3                   & 20.4 ± 0.1                     & 37.1 ± 0.3                  & 42.6 ± 0.2                   & 42.0 ± 0.1                     \\
CIDD & \underline{51.3 ± 0.4}         & \underline{50.2 ± 0.3}          & \underline{21.2 ± 0.3}            & \underline{50.4 ± 0.5}         & \underline{51.3 ± 0.2}          & \underline{43.2 ± 0.2}            \\ 
CIDD+NRR & \textbf{66.7 ± 0.4}         & \textbf{55.7 ± 0.2}          & \textbf{25.6 ± 0.2}            & \textbf{65.1 ± 0.3}         & \textbf{58.3 ± 0.2}          & \textbf{51.3 ± 1.0}            \\ \bottomrule

\end{tabular}
\end{adjustbox}
\caption{Comparison between the baseline RDED \cite{rded} and our framework using only the CIDD modules (CIDD) and with Non-Critical Region Refinement (NRR) (CIDD+NRR). All methods are evaluated with IPC = 10.}
\label{tab:ab-cidd}
\end{table}

\noindent
\textbf{Different Label Compression Comparison.} To demonstrate the benefits of our distance-based representation (DBR), we conducted experiments to compare it with various label compression methods. The results show that, while requiring storage of only two distance values per instance (achieving 5× compression on ImageNette and 500× on ImageNet1k compared to soft-labels), our method provides comparable performance. Notably, in all cases with IPC set to 1, our DBR method outperforms the state-of-the-art RDED with soft-labels. When compared to compact labels (which use only two elements of soft-labels), our method achieves significant improvements across all scenarios. These findings highlight the effectiveness of distance-based representation in this task.

\noindent
\textbf{Cross-architecture Generalization.} To ensure the generalization capability of our distilled datasets, it is crucial to evaluate their performance across a range of neural architectures not involved in the dataset distillation process. The results in Table \ref{tab:cross} demonstrate the robustness and generalization of our method across all cross-comparisons. This can be attributed to the advantage of selecting the critical key patches in CIDD and the effectiveness of non-critical image refinement.

\subsection{Ablation Study}
\noindent
\textbf{Effectiveness of Critical-based Initial Data Discovery.} To verify the benefits of our Critical-based Initial Data Discovery (CIDD) in Section \ref{sec:ids}, we conducted experiments using only the data generated by CIDD to train the student model, following the same settings as RDED \cite{rded}. The results in Table \ref{tab:ab-cidd} demonstrate that our CIDD achieves better performance than RDED in all cases, indicating the effectiveness of the module.

\noindent
\textbf{Effectiveness of Non-Critical Region Refinement.} Table \ref{tab:ab-cidd} shows the results of our model with NRR (CIDD+NRR) and without NRR (CIDD). The results demonstrate that the NRR method significantly improves model performance. For example, on CIFAR-10 using ResNet18, the addition of NRR leads to an approximate 15\% performance increase compared to when NRR is not used. This clearly indicates the effectiveness of incorporating the Non-Critical Region Refinement module.

\noindent
\textbf{Effectiveness of Label Refinement ($\mathcal{L}_{lr}$).} To further validate the benefits of $\mathcal{L}_{lr}$ in Eq. \ref{eq:lcnew}, we conducted experiments to determine whether this term enhances model learning without relying on soft-label information. As shown in Table \ref{tab:dr}, incorporating Label Refinement NRR-DD (DBR) consistently improves performance over without using Label Refinement NRR-DD (DBR) across all comparisons, demonstrating the effectiveness of this term.
\section{Limitation and Future Works}

A possible limitation of our work is its reliance on the quality of Class Activation Mapping (CAM) for identifying critical and non-critical regions. While CAM provides useful insights, its performance can be sensitive to the model's initial training, potentially affecting the accuracy of the identified regions. Therefore, a promising direction for future work is exploring how to connect the training of pretrained models to generate better CAMs.

Additionally, while the Distance-Based Representative (DBR) technique reduces memory requirements, it may not fully address the trade-off between memory efficiency and the preservation of fine-grained details for certain complex tasks. This emphasizes the need for further research to develop enhanced solutions for this issue, which we plan to explore in our future work.

\section{Conclusion}
In this paper, we introduced the Non-Critical Region Refinement Dataset Distillation (NRR-DD) method to address the limitations of current dataset distillation techniques. Our approach comprises three key stages: Critical-based Initial Data Discovery (CIDD) to capture instance-specific details, Non-Critical Region Refinement (NRR) to balance critical and non-critical regions using Class Activation Mapping, and relabeling to transfer knowledge effectively. Additionally, we proposed the Distance-Based Representative (DBR) technique, which eliminates the need for memory-intensive soft label storage by using a distance-based measure, significantly reducing memory requirements. By combining DBR with NRR, our method generates compact and efficient datasets, achieving state-of-the-art performance across small- and large-scale datasets. Experimental results confirm that our approach not only enhances dataset representativeness but also minimizes training complexity, making it highly suitable for various training environments. This work advances dataset distillation by capturing critical features and optimizing storage, paving the way for efficient, scalable data solutions.

\section*{Acknowledgements}
This work was supported by ARC DP23 grant DP230101176 and by the Air Force Office of Scientific Research under award number FA2386-23-1-4044.

{\small
\bibliographystyle{ieee_fullname}
\bibliography{egbib}
}

\appendix
\section{Training Details}

For all experiments conducted in this study, we fixed the following hyperparameters:

\begin{itemize} 
\item Scale factor for batch normalization $\alpha_{bn} = 10$ (as follows to \cite{sre2l})
\item CAM matrix's upper threshold $\epsilon = 0.5$ 
\item Embedding radius $r = 0.4$ 
\item Scale factors for distance-based representation $\alpha_{dbr} = 1$ and $\alpha_{lr} = 1$ 
\end{itemize}

These values were selected through careful tuning based on prior work and were maintained consistently across all experiments to ensure comparability and minimize the impact of hyperparameter variability. By fixing these hyperparameters, we aim to provide a fair and reproducible assessment of our methods across different datasets and experimental conditions.

\subsection{Non-Critical Region Refinement Phase.} During this phase, we utilized the following settings:

\begin{itemize} 
\item Optimizer: Adam 
\item Learning rate: 0.05 
\item Betas: $\beta_1 = 0.5$, $\beta_2 = 0.9$ 
\item Batch size: 100 
\item Iterations: 2000 
\end{itemize}

This phase is designed to refine the synthetic data, particularly in non-critical regions of the feature space, enhancing the model's ability to generalize for subsequent tasks.

\subsection{Knowledge Transfer and Post-Evaluation Phase.} In this phase, we applied the following parameters:

\begin{itemize} 
\item Optimizer: AdamW (incorporating weight decay for better generalization) 
\item Learning rate: 1e-3 
\item Batch size: 100 
\item Training epochs: 300 
\item Learning rate scheduler: Smoothing LR 
\end{itemize}

To augment the synthetic data and improve robustness, we employed the following data augmentation techniques:

\begin{itemize} 
\item Two RandAugment transformations 
\item RandomResizeCrop 
\item RandomHorizontalFlip 
\end{itemize}

These augmentations, as detailed in \cite{rded} and \cite{sre2l}, help prevent overfitting by introducing variability into the data, thereby enhancing the model's generalization capabilities.

All of these settings were consistently applied across all datasets to ensure fair comparison and reliable evaluation of the model’s performance under varying experimental conditions.

\section{Parameter Sensitivity Analysis}

\subsection{Threshold $\epsilon$.} In this section, we examine the impact of different threshold values ($\epsilon$) on model performance across two datasets, ImageNette and CIFAR100, for both IPC 10 and IPC 50 classification tasks. The results presented in Table \ref{tab:ps:e} show that a threshold of 0.5 consistently delivers the best performance across all settings. Specifically, for ImageNette (IPC 10), the accuracy reaches 66.2\%, and for ImageNette (IPC 50), it peaks at 85.6\%. Similarly, for CIFAR100 (IPC 10), the highest accuracy is 62.7\%, while CIFAR100 (IPC 50) achieves 67.1\%. This optimal performance at $\epsilon = 0.5$ can be attributed to a balanced trade-off, where this threshold maintains an effective level of value retention and updates without excessive loss or staleness.

\begin{table}[h]
\begin{adjustbox}{width=\linewidth}
\begin{tabular}{@{}lccccccc@{}}
\toprule
Thresold $\epsilon$               & 0.1   & 0.3   & 0.4   & 0.5           & 0.6   & 0.7   & 0.9   \\ \midrule
ImageNette   (IPC 10) & 63.15 & 65.35 & 65.59 & \textbf{66.2} & 65.87 & 65.4  & 65.35 \\
ImageNette   (IPC 50) & 82.89 & 84.66 & 85.01 & \textbf{85.6} & 85.08 & 84.7  & 84.31 \\
CIFAR100  (IPC 10)     & 60.14 & 61.56 & 62.28 & \textbf{62.7} & 61.29 & 61.69 & 61.14 \\
CIFAR100  (IPC 50)     & 66.58 & 65.5  & 66.46 & \textbf{67.1} & 66.26 & 66.7  & 66.15 \\ \bottomrule
\end{tabular}
\end{adjustbox}
\caption{Comparison of model performance across different threshold values ($\epsilon$) on the ImageNette and CIFAR100 datasets for IPC 10 and IPC 50 classification tasks. The best performance is achieved at $\epsilon = 0.5$ for both datasets and tasks.}

\label{tab:ps:e}
\end{table}

\subsection{Radius $r$.} In this section, we investigate the impact of different radius values on model performance across the ImageNette and CIFAR10 datasets for both IPC 10 and IPC 50 classification tasks. The results in Table \ref{tab:ps:r} reveal that a radius of 0.4 consistently provides the best performance across all settings. Specifically, for ImageNette (IPC 10), the accuracy peaks at 66.2\%, and for ImageNette (IPC 50), it reaches 85.6\%. Similarly, for CIFAR10 (IPC 10), the highest accuracy is 62.7\%, while CIFAR10 (IPC 50) achieves 67.1\%. This suggests that a radius of 0.4 strikes an optimal balance, yielding the highest accuracy without causing overfitting or excessive loss.

\begin{table}[h]
\begin{adjustbox}{width=\linewidth}
\begin{tabular}{@{}lccccccc@{}}
\toprule
radius                & 0.1   & 0.2   & 0.3   & 0.4           & 0.5   & 0.6   & 0.7   \\ \midrule
ImageNette   (IPC 10) & 64.29 & 65.24 & 65.24 & \textbf{66.2} & 64.92 & 65.6  & 64.91 \\
ImageNette   (IPC 50) & 84.17 & 83.41 & 84.21 & \textbf{85.6} & 84.14 & 85.24 & 85.05 \\
CIFAR10  (IPC 10)     & 61.02 & 61.21 & 61.97 & \textbf{62.7} & 61.18 & 61.16 & 59.56 \\
CIFAR10  (IPC 50)     & 65    & 65.1  & 66.02 & \textbf{67.1} & 65.74 & 66.57 & 65.74 \\ \bottomrule
\end{tabular}
\end{adjustbox}
\caption{Comparison of model performance across different radius values on the ImageNette and CIFAR10 datasets for IPC 10 and IPC 50 classification tasks. A radius of 0.4 consistently yields the best performance across all settings.}
\label{tab:ps:r}
\end{table}

\subsection{Scale Factor $\alpha_{dbr}$ and $\alpha_{lr}$.} In this section, we analyze the effect of two hyperparameters, $\alpha_{dbr}$ and $\alpha_{lr}$, on model performance across different configurations. The results, as shown in Table \ref{tab:ps:dbr}, indicate that the best performance is achieved when $\alpha_{dbr} = 1$ and $\alpha_{lr} = 1$, yielding an accuracy of \textbf{66.2}\%. This combination outperforms other configurations, particularly for higher values of $\alpha_{lr}$, where the accuracy tends to decrease. These findings suggest that the balance between these two parameters plays a critical role in optimizing the model’s performance.

\begin{table}[h]
\centering
\begin{adjustbox}{width=0.9\linewidth}
\begin{tabular}{|l|c|c|c|c|c|}
\hline
\diagbox[width=\dimexpr \textwidth/12+2\tabcolsep\relax, height=0.6cm]{ $\alpha_{dbr}$ }{$\alpha_{lr}$} & 0.2  & 0.5  & 1             & 2     & 5    \\ \hline
0.2  & 64.2 & 64.5 & 64.4          & 64.4  & 62.7 \\ \hline
0.5  & 63.7 & 65.0 & 65.6          & 65.0  & 62.6 \\ \hline
1    & 63.8 & 65.8 & \textbf{66.2} & 65..9 & 63.2 \\ \hline
2    & 64.7 & 66.1 & 65.6          & 64.3  & 62.8 \\ \hline
5    & 64.5 & 63.6 & 63.8          & 63.9  & 63.0 \\ \hline
\end{tabular}
\end{adjustbox}
\caption{Performance analysis of the model with varying values of $\alpha_{dbr}$ and $\alpha_{lr}$. The highest accuracy of \textbf{66.2}\% is achieved when $\alpha_{dbr} = 1$ and $\alpha_{lr} = 1$.}
\label{tab:ps:dbr}
\end{table}

\subsection{Ablation Study on $L_{bn}$:} Inspired by your recommendations, we conducted additional experiments in the Table \ref{tab:lbn}. The results show that: (1) With or without $L_{bn}$, our method outperforms RDED; (2) $L_{bn}$ is important and helps improve model performance.

\begin{table}[h]
\centering
\begin{adjustbox}{width=1\linewidth}
\begin{tabular}{@{}lcccc@{}}
\toprule
            & \multicolumn{1}{l}{CIFAR10} & \multicolumn{1}{l}{CIFAR100} & \multicolumn{1}{l}{ImageNette} & \multicolumn{1}{l}{ImageNet1k} \\ \midrule
RDED        & 37.1 ± 0.3                  & 42.6 ± 0.2                   & 61.4 ± 0.4                     & 42.0 ± 0.1                     \\
NRR-DD Without $L_{bn}$ & 68.3 ± 0.4                  & 60.1 ± 0.2                   & 65.1 ± 0.5                     & 44.8 ± 0.2                     \\
NRR-DD With $L_{bn}$    & 72.2 ± 0.4                  & 62.7 ± 0.2                   & 66.2 ± 0.6                     & 46.1 ± 0.2                     \\ \bottomrule
\end{tabular}
\end{adjustbox}
\caption{Ablation Study for $L_{bn}$ with 10 IPC.}
\label{tab:lbn}
\end{table}

\section{Choosing Lowest Confident Score or Highest Confident Score.} 

\begin{table*}[hpt]
\centering
\begin{adjustbox}{width=0.9\linewidth}
\begin{tabular}{@{}lcccccc@{}}
\toprule
                           & \multicolumn{3}{c}{ConvNet}                                                                 & \multicolumn{3}{c}{Resnet18}                                                                \\ \midrule
                           & \multicolumn{1}{l}{CIFAR10} & \multicolumn{1}{l}{CIFAR100} & \multicolumn{1}{l}{ImageNet1k} & \multicolumn{1}{l}{CIFAR10} & \multicolumn{1}{l}{CIFAR100} & \multicolumn{1}{l}{ImageNet1k} \\
Highest-Score   (Direct)   & 50.2 ± 0.3                  & 48.1 ± 0.3                   & 20.4 ± 0.1                     & 37.1 ± 0.3                  & 42.6 ± 0.2                   & 42.0 ± 0.1                     \\
Lowest-Score   (Direct)    & \textbf{51.3 ± 0.4}         & \textbf{50.2 ± 0.3}          & \textbf{21.2 ± 0.3}            & \textbf{50.4 ± 0.5}         & \textbf{51.3 ± 0.2}          & \textbf{43.2 ± 0.2}            \\
Highest-Score   (Refining) & 64.8± 0.4                   & 53.1 ± 0.3                   & 24.1 ± 0.3                     & 63.7 ± 0.2                  & 57.1 ± 0.3                   & 49.9 ± 1.1                     \\
Lowest-Score   (Refining)  & \textbf{66.7 ± 0.4}         & \textbf{55.7 ± 0.2}          & \textbf{25.6 ± 0.2}            & \textbf{65.1 ± 0.3}         & \textbf{58.3 ± 0.2}          & \textbf{51.3 ± 1.0}            \\ \bottomrule
\end{tabular}
\end{adjustbox}
\caption{Comparison of the performance using Lowest Confident Score (LCS) versus Highest Confident Score (HCS) for selecting initial patches. The results show that LCS outperforms HCS in all scenarios, both in direct training and after applying Non-Critical Refinement (Refining), highlighting the advantages of LCS in improving model performance.}
\label{tab:ps:lshs}
\end{table*}

To further highlight the benefits of using the Lowest Confident Score (LCS) over the Highest Confident Score (HCS) for selecting initial patches, we compare two experimental scenarios: (1) training the combined images of these patches directly (referred to as Direct) to evaluate the raw image quality, and (2) refining the images through our Non-Critical Refinement (Referred to as Refining) to assess the improved versions. The results, presented in Table \ref{tab:ps:lshs}, clearly indicate that the Lowest Confident Score consistently yields better performance than the Highest Confident Score across all evaluated settings. These observations can be attributed to two main factors:

\begin{itemize}
    \item \textbf{Hard-to-Learn Samples:} The Lowest Confident Score tends to identify harder-to-learn, more challenging samples that are often neglected by simpler approaches. These samples provide critical information that enhances the model’s ability to generalize, which leads to improved performance across different datasets. Additionally, by selecting these harder samples, the model still focus on instance-specific features, which is the most important features in large-scale dataset distillation.
    
    \item \textbf{Optimization Flexibility with CE Loss:} When using the teacher’s classification loss (CE Loss) on synthetic data, the Lowest Confident Score offers more flexibility for parameter updates compared to the Highest Confident Score. This is because patches with the Highest Confident Score typically have a CE Loss close to zero in many cases, indicating that these samples are already well-classified and do not require significant adjustments. In contrast, the Lowest Confident Score corresponds to harder-to-learn patches, which are more likely to lead to meaningful updates during training. By focusing on these difficult samples, the model has more opportunity for optimization, thus facilitating better overall performance.
\end{itemize}

Thus, the results demonstrate that using the Lowest Confident Score for selecting patches not only improves the model’s performance but also makes the training process more effective. By focusing on hard-to-learn samples, LCS enables the model to update parameters more significantly, yielding better performance both in direct training and after refinement. The effectiveness of LCS in this context underscores its importance as a selection criterion for initial patches.

\section{Further Discussion}

\noindent
\textbf{\textit{Balancing instance-level vs class-level features:}} Thank you for your comment. We’ve added a discussion in the revised paper, explaining the effect of instance- and class-level information by using different value of \( \epsilon \) in Eq. 4 (higher \( \epsilon \) emphasizes lower instance-level and higher class-level information). The table below shows optimal performance at \( \epsilon = 0.5 \), balancing both levels. Qualitative images with different \( \epsilon \) values are also included for illustration.
\raggedbottom
\begin{figure}[h]
\centering
\includegraphics[width=1\linewidth]{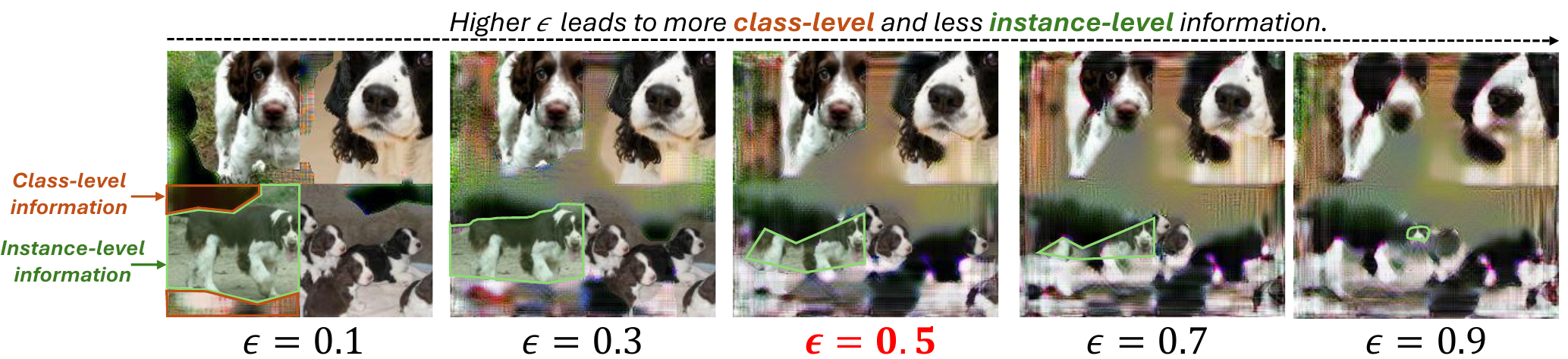} 
\label{fig:visual1}
\caption{Visualization with various value of $\epsilon$.}
\end{figure}

\begin{table}[h]
\centering
\begin{adjustbox}{width=1\linewidth}
\begin{tabular}{@{}lccccccc@{}}
\toprule
Thresold $\epsilon$               & 0.1   & 0.3   & 0.4   & {\color{red}\textbf{0.5 } }         & 0.6   & 0.7   & 0.9   \\ \midrule
CIFAR100  (IPC 10)     & 60.14 & 61.56 & 62.28 & {\color{red}\textbf{62.7}} & 61.29 & 61.69 & 61.14 \\
CIFAR100  (IPC 50)     & 66.58 & 65.5  & 66.46 & {\color{red}\textbf{67.1}} & 66.26 & 66.7  & 66.15 \\ \bottomrule
\end{tabular}
\end{adjustbox}
\caption{Accuracies with various value of $\epsilon$.}
\label{tab:eff}
\end{table}

\noindent
\textbf{\textit{Efficiency test:}} The experiment in Table \ref{tab:eff} shows our method has lower time cost and similar memory to SRe2L. However, due to image refinement, both time and peak memory are higher than RDED.
\begin{table}[h]
\centering
\begin{adjustbox}{width=1 \linewidth}
\begin{tabular}{@{}lllllll@{}}
\toprule
                 & \multicolumn{3}{c}{Resnet18} & \multicolumn{3}{c}{MobileNet-V2} \\ \midrule
Architecture     & SRe2L     & RDED   & Our     & SRe2L      & RDED     & Our      \\
Time Cost (ms)   & 2113.23   & 39.89  & 520.65  & 3783.16    & 64.97    & 989.46   \\
Peak Memory (GB) & 9.14      & 1.57   & 9.14    & 12.93      & 2.35     & 12.93    \\ \bottomrule
\end{tabular}
\end{adjustbox}
\caption{Efficiency test in generating 100 images in ImageNet1k.}
\label{tab:eff}
\end{table}

\noindent
\textbf{\textit{Neural architecture search (NAS):}} Due to time constraints, we follow the setting in the DM paper and run experiement in Table \ref{tab:nas}. Results show that our synthetic data achieves better performance than previous methods.

\begin{table}[h]
\centering
\begin{adjustbox}{width=1\linewidth}
\begin{tabular}{@{}lcccccc@{}}
\toprule
\multicolumn{1}{c}{} & Random & DSA   & DM    & RDED  & Our   & Whole dataset \\ \midrule
Performance (\%)     & 84.0     & 82.6  & 84.3  & 84.6  & 84.9  & 85.9          \\
Correlation          & -0.04  & 0.68  & 0.76  & 0.78  & 0.80  & 1.00             \\
Time cost (min)      & 142.6  & 142.6 & 142.6 & 142.6 & 142.6 & 3580.2        \\ \bottomrule
\end{tabular}
\end{adjustbox}
\caption{NAS on CIFAR-10 (50 IPC) searching for 720 ConvNets.}
\label{tab:nas}
\end{table}

\noindent
\textbf{\textit{MMT's Initialization:}} Table \ref{tab:init} shows results with MMT-initialized images. Our method only yields slight improvements, as these images already reside in high-confidence regions and contain class-general information. The near-zero $L_C$ leads to minimal refinement, and their limited fine-grained details keep performance relatively low.

\noindent
\textbf{\textit{Real Data Initialization:}} On the other hand, real data initialization performs well, achieving SOTA results, though slightly lower than our method. This is due to the strong instance-specific information in real images, with our refining method adding class-general features. 

\begin{table}[h]
\centering
\begin{adjustbox}{width=\linewidth}

\begin{tabular}{@{}lllccc@{}}
\toprule
                  & MMT (Baseline) & \multicolumn{1}{c}{MMT's Init} & Real Image & RDED's Init & NRR-DD's Init \\ \midrule
CIFAR100 (10 IPC) & 40.1 ± 0.4     & 40.8 ± 0.6                     & 53.5 ± 0.3 & 53.8 ± 0.3  & 55.7 ± 0.2    \\
CIFAR100 (50 IPC) & 47.7 ± 0.2     & 48.0 ± 0.3                     & 59.5 ± 0.1 & 59.6 ± 0.1  & 61.1 ± 0.1    \\               
 \bottomrule
\end{tabular}
\end{adjustbox}
\caption{Comparing various initializations.}
\label{tab:init}
\end{table}

\end{document}